\documentclass[journal]{IEEEtran}
\usepackage{stfloats} 
\usepackage{color}
\usepackage{tabularx}
\usepackage{booktabs}
\usepackage{graphicx} 
\usepackage{hyperref}
\usepackage{amsmath}
\usepackage{amssymb}
\usepackage{dsfont}
\usepackage{float} %设置图片浮动位置的宏包
\usepackage{subfig} %插入多图时用子图显示的宏包
\usepackage{graphicx, subcaption}

\usepackage{algorithmic}
\usepackage{algorithm}

\begin{document}

\title{Incremental Pseudo-Labeling for Black-Box Unsupervised Domain Adaptation}

\author{Yawen~Zou,~\IEEEmembership{University of Fukui,}
        Chunzhi~Gu,~\IEEEmembership{Toyohashi University of Technology,}
        Jun~Yu,~\IEEEmembership{Niigata University,}
        Shangce~Gao,~\IEEEmembership{University of Toyama,}
        and~Chao~Zhang,~\IEEEmembership{University of Toyama}
%\thanks{M. Shell was with the Department
%of Electrical and Computer Engineering, Georgia Institute of Technology, Atlanta,
}

\maketitle
\begin{abstract}
Black-Box unsupervised domain adaptation (BBUDA) learns knowledge only with the prediction of target data from the source model without access to the source data and source model, which attempts to alleviate concerns about the privacy and security of data. However, incorrect pseudo-labels are prevalent in the prediction generated by the source model due to the cross-domain discrepancy, which may substantially degrade the performance of the target model. To address this problem, we propose a novel approach that incrementally selects high-confidence pseudo-labels to improve the generalization ability of the target model. Specifically, we first generate pseudo-labels using a source model and train a crude target model by a vanilla BBUDA method. Second, we iteratively select high-confidence data from the low-confidence data pool by thresholding the softmax probabilities, prototype labels, and intra-class similarity. Then, we iteratively train a stronger target network based on the crude target model to correct the wrongly labeled samples to improve the accuracy of the pseudo-label. Experimental results demonstrate that the proposed method achieves state-of-the-art black-box unsupervised domain adaptation performance on three benchmark datasets.
\end{abstract}

\begin{IEEEkeywords}
Black-Box Unsupervised Domain Adaptation (BBUDA), Pseudo-Labeling, Intra-Class Similarity.
\end{IEEEkeywords}

\IEEEpeerreviewmaketitle

\section{Introduction}
Deep neural networks have shown tremendous growth with large-scale, manually labeled data in the past few decades. However, manual annotation of data is highly expensive and time-intensive. Unsupervised domain adaptation (UDA) \cite{wang2018deep} methods, which transfer the knowledge learned from previously labeled datasets to unlabeled target datasets, have been devised to tackle this problem in terms of domain adaptation. Specifically, UDA has been extensively researched in recent years in various fields, including object recognition \cite{ganin2016domain,liang2018aggregating,tzeng2017adversarial} and semantic segmentation \cite{zhang2017curriculum,tsai2018learning}.

Most existing UDA methods involve access to source data, potentially raising concerns regarding data privacy and security, and the process of domain adaptation becomes inefficient when the source data is too large. Therefore, source-free UDA \cite{yang2020unsupervised, hou2020source,liang2020we} was introduced to adapt the source model to the target domain without accessing source data. In particular, the method that exclusively employs the source model without recourse to source data is termed white-box unsupervised domain adaptation (WBUDA) \cite{zhang2021unsupervised}. WBUDA fine-tunes the source model to fit the target domain with unlabeled target data. Yang et al. \cite{yang2020unsupervised} introduced an additional classifier that applies the classifier of the source model as an initial state to achieve target feature alignment with the corresponding prototypes of the source classifier. Hou et al. \cite{hou2020source} transfer the style of the target domain to that of unseen source data stored in the source classifier's Batch Normalization (BN) layer before being fed into it. 

\begin{figure}[tb] %H为当前位置，!htb为忽略美学标准，htbp为浮动图形
\centering %图片居中
\includegraphics[width=0.5\textwidth]{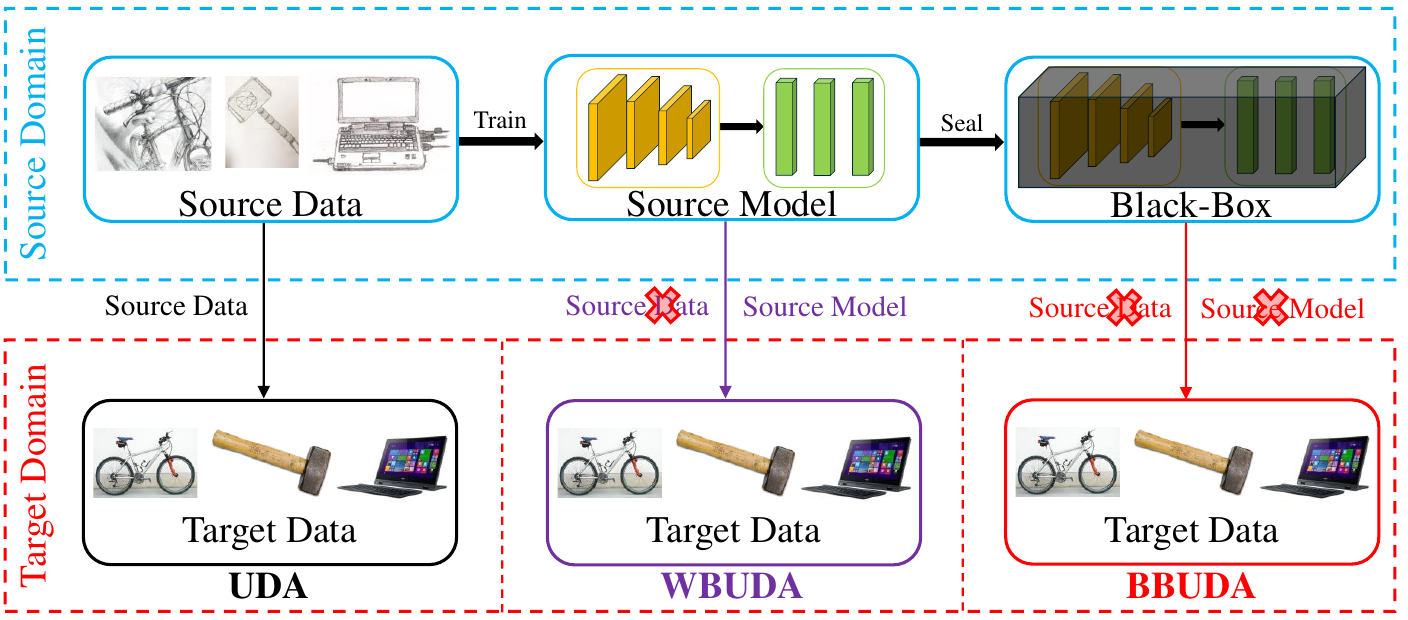} %插入图片，[]中设置图片大小，{}中是图片文件名
\caption{Illustration of different UDA settings. Source data is required in traditional UDA, whereas WBUDA is source data free but requires access to the source model. BBUDA requires the prediction of target data from the black box without access to both the source data and the source model.
} %最终文档中希望显示的图片标题
\label{Figure.1} %用于文内引用的标签
\end{figure}

On the other hand, the source model could be inaccessible due to commercial or safety constraints in practice. In a realistic setting, most commercial models only provide predictions of unlabeled target data via the cloud API service, which can help commercialize the model and avoid white-box attacks \cite{thys2019fooling}. Essentially, the source model acts as a black box. In such a case, the task of black-box Unsupervised Domain Adaptation (BBUDA) \cite{liang2022dine,zhang2023black,peng2023rain,zhang2021unsupervised}, which only requires the prediction of target data from the source model without access to the source data and information on the source model, has been proposed to alleviate these concerns. Compared with UDA and WBUDA as shown in Fig. \ref{Figure.1}, BBUDA is the most challenging and safe domain adaptation method without utilizing any information about the source domain.

Although BBUDA methods have achieved remarkable progress, one issue is far from being explored. Incorrect pseudo-labels are prevalent in the initial prediction generated by the source model due to the cross-domain discrepancy, which has a noteworthy impact on the performance of the target model and has been sparsely studied. It should be noted that in most methods, the pseudo-labels are directly and fully exploited from the source model without considering the potential influence of incorrect labels \cite{liang2022dine,peng2023rain,shi2023source}. Consequently, the resulting errors introduced by pseudo-labels can accumulate, damaging the final performance.

To alleviate the above issue, in this study, we propose incrementally increasing high-confidence data to improve the reliability of the pseudo-labels and thus further contribute to boosting BBUDA. The core idea of our algorithm is to maintain the high-confidence pseudo-labels via thresholding of softmax probabilities, prototype labels (pseudo-labels via nearest centroids) \cite{liang2020we}, and intra-class similarity \cite{wang2023improving} in each optimization iteration to train a better target network to correct the low-confidence pseudo-labels. As stated in  \cite{hendrycks2016baseline}, correctly classified target instances generally have greater maximum softmax probabilities compared to misclassified instances. Hence, thresholding at the highest softmax probability can be used to maintain the true predictions of neural networks in classification tasks. However, the target model is always prone to overconfident predictions compared to the source model. Hence, in this paper, we only use this method to keep high-confidence data directly from the source model's pseudo-labels via a high threshold in the warm-up process. Also, during the incremental process, we set a low threshold to reject the low-confidence pseudo-labels. In addition, inspired by prior works \cite{liang2020we,liang2022dine}, our method generates the prototype labels via the nearest centroids. However, due to the BBUDA setting, features of target data generated by the source model are inaccessible. Thus, we introduce the teacher-student knowledge distillation mechanism to learn structural features to generate prototype labels. 

Specifically, unlike previous studies that simply take the closest centroid as the prototype
label without selection, we only utilize prototype labels that satisfy the two conditions: 1) a significant difference of distance occurs between the closest centroid and the second closest centroid, and 2) pseudo-label consistency is maintained between the prototype label and the softmax output. Meanwhile, the intra-class similarity \cite{wang2023improving} of pseudo-labels is used to discard data with low similarity scores because candidates with high scores are more likely to belong to the same class. The finally selected high-confidence pseudo-labeled samples are fed to the crude model generated by the vanilla BBUDA method to improve the performance. %In addition, we introduce a warm-up process in which only the high-confidence data selected from the prediction of the black-box source model are used to train the crude target model.
The model is then iteratively trained resulting in incrementally increased high-confidence pseudo-labels. By introducing the incremental pseudo-labeling mechanism, our method relaxes the requirement of obtaining a strong target network with high-confidence data in the initial training phase. Our method can be applied to any BBUDA method to utilize high-confidence samples to reduce the impact of wrongly labeled target samples and thus improve the generalization capabilities. The contributions of this study are threefold.

\begin{itemize}
    \item We propose an incremental pseudo-labeling method for BBUDA by iteratively increasing the number of high-confidence samples and iteratively training a stronger target network to correct the low-confidence pseudo-labels. 
    \item We introduce a warm-up process at the initial training stage, in which a teacher-student knowledge distillation mechanism is incorporated to have easier access to the target features. 
% Second, we propose a high-confidence selection method that keeps the high-quality pseudo-labels and rejects low-quality pseudo-labels. It explores pseudo-labels containing incorrect predictions and assigns them to a high-confidence or low-confidence data pool.   
    \item We performed extensive experiments and ablation studies to demonstrate the effectiveness of our method against prior BBUDA models in terms of classification accuracy.
% Third, our method introduces a student network to distill structure knowledge from the target predictions of the black-box source model, which can be used to get the target feature to generate the prototype labels. In addition, it can be extended to other BBUDA methods, and experimental results on several benchmark datasets show that it can improve the performance of models.

\end{itemize}

\section{Related work}

\noindent\textbf{Unsupervised Domain Adaptation (UDA)}. UDA aims to transfer knowledge in the labeled source domain to the unlabeled target domain to solve tasks similar to those in the source domain. Over the last few years, UDA has been broadly applied in object recognition \cite{ganin2016domain,liang2018aggregating}, object detection \cite{chen2018domain,khodabandeh2019robust}, image generation \cite{liu2022deep}, and semantic segmentation \cite{tsai2018learning} following the rapid growth of neural network research. The existing UDA methods include statistical moment matching \cite{long2015fully,sun2016return,tzeng2014deep}, feature adversarial learning \cite{ganin2016domain,liu2021adversarial,hoffman2018cycada}, and self-training \cite{zou2019confidence,zou2018unsupervised}. Statistic moment matching uses various discrepancy metrics, such as maximum mean discrepancy (MMD) \cite{gretton2006kernel}, correlation alignment (CORAL) \cite{sun2016return}, contrastive domain discrepancy (CDD) \cite{kang2019contrastive} to minimize the discrepancy between the source domain and target domain in a latent feature space. Feature adversarial learning \cite{ganin2016domain,tzeng2017adversarial} utilizes adversarial training to extract invariant domain features, which are utilized to predict the source sample labels but indiscriminate to the change in domains. Another popular paradigm \cite{liu2021generative} leverages the idea of self-training and introduces a practical Bayesian uncertainty mask to filter the pseudo-labels. However, it is not always practical and unsafe in many cases to exploit knowledge from labeled source domain data for these methods, which may raise concerns about data leaks.

\noindent\textbf{Black-Box Unsupervised Domain Adaptation (BBUDA)}. BBUDA learns knowledge from target predictions provided by trained source model rather than from source data and source model for domain adaptation \cite{liang2022dine}. BBUDA is significantly more secure than the white-box source model and lowers privacy concerns for source data and models. Because there is no risk of data leakage, BBUDA is regarded as a desirable UDA configuration. A few recent attempts \cite{liang2022dine,zhang2023black,peng2023rain,zhang2021unsupervised,shi2023source} have addressed the BBUDA problem. Liang et al. \cite{liang2022dine} distilled the knowledge from the trained source network sealed as an application programming Interface (API) to train a target network, then fine-tuned the distilled network to improve the performance target network. Peng et al. \cite{peng2023rain} combined a new data augmentation called Phase MixUp and a novel regularization technique at the subnetwork level during domain adaptation. Zhang et al. \cite{zhang2023black} divided the generated features into sensory, short-term, and long-term memory to maintain useful and representative information. Zhang et al.  \cite{zhang2021unsupervised} utilized a noisy label learning approach to assign distinct sampling thresholds for different categories to address the unbalanced label noise.
Shi et al. \cite{shi2023source} exploited supervision information between the source model and public third-party data as a bridge to mitigate the gap between source data and target data by Distributionally Adversarial Training (DAT). In contrast to these studies \cite{liang2022dine,peng2023rain,zhang2023black,shi2023source}, our method incrementally selects high-confidence data to iteratively train a better target network for correcting low-confidence labels instead of resorting directly to the initial pseudo-labels. It also serves as a unified technique for UDA methods to improve generality.

\begin{figure*}[t] %H为当前位置，!htb为忽略美学标准，htbp为浮动图形
\centering %图片居中
\includegraphics[width=0.8\textwidth]{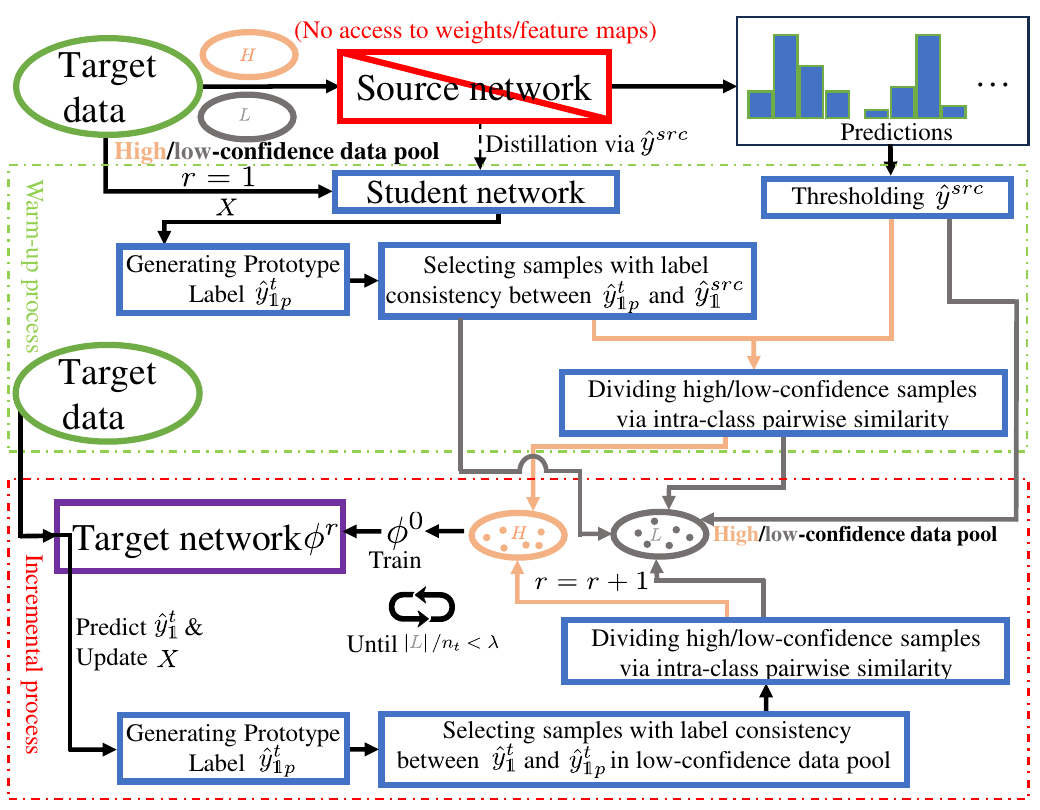} %插入图片，[]中设置图片大小，{}中是图片文件名
\caption{Overview of the proposed framework, consisting of warm-up and incremental processes. For the warm-up process, a trained source network provides the predictions of target data (soft label $\hat{y}^{src}$ and hard label $\hat{y}^{src}_{\mathds{1}}$). A student network is distilled from the source model to extract the features $X$ of the target data and obtain prototype labels $\hat{y}^{t}_{\mathds{1}p}$. Then, high-confidence samples are selected and passed to the incremental process. For the incremental process, the $r$-th cycle's prototype labels and features $X$ are updated by the target network $\phi^{r} $ instead of the student work. Then, high-confidence data are selected from the low-confidence data pool in each iteration via pseudo-label consistency and intra-class similarity until the percentage of low-confidence data is less than $\lambda$.
} %最终文档中希望显示的图片标题
\label{Figure.3} %用于文内引用的标签
\end{figure*}

\noindent\textbf{Pseudo-Labeling (PL)}. 
Pseudo-labeling is a widely introduced approach for distilling knowledge from the source domain in UDA. It is generally assumed that PL \cite{lee2013pseudo} selects the class with the highest prediction probability obtained from the trained model as true labels. Early PL methods directly learn knowledge from pseudo-labels of unlabeled target samples without selection. Long et al. \cite{long2013transfer} use hard pseudo target labels with the source data labels to measure the modified MMD distance between domains. Gebru et al. \cite{tzeng2015simultaneous} apply soft pseudo-labels rather than hard pseudo-labels to learn the categories relationships between the source domain and target domain. However, some low-confidence predictions obtained from the source model are often inaccurate, which inevitably causes an error accumulation issue during the training process. Currently, a new variant of the PL method has been devised to pick pseudo-labels with high reliability to alleviate mislabeling errors \cite{zheng2021rectifying,chen2019progressive,liang2021source,wang2023improving}. Zheng and Yang \cite{zheng2021rectifying} put forward a method to directly estimate the uncertainty of pseudo-labels through the variance of prediction between two classiﬁers trained with different layers. Chen et al. \cite{chen2019progressive} select reliable pseudo-labels via the cross-domain similarity measurements of representations to align the class distributions between the source domain and target domain. Liang et al. \cite{liang2021source} divide the target data of each class into ``easy'' and ``hard'' samples according to the entropy values of pseudo-labels. Wang et al. \cite{wang2021source} select positive pseudo-labels with high confidence scores by keeping the top $K$ predictions within per class. These methods achieve good generalization ability of the model. Distinct from the above techniques, our method introduces prototype labels and intra-class similarity to select high-confidence data, which can generate more reliable pseudo-labels.

\section{Method}
Our proposed method aims to tackle black-box unsupervised domain adaptation with only pseudo-labels from a pre-trained source model without access to the source model and source data. 
Specifically, we incrementally improved the accuracy of the pseudo-labels in the target domain using the following three selection methods: thresholding of softmax probabilities, prototype labels (pseudo-labels via nearest centroids), and intra-class similarity. An overview of the proposed method is shown in Fig. \ref{Figure.3}. The whole procedure can be divided into two parts: the warm-up process and the incremental process. A source model pre-trained with labeled source data is given to provide the initial predictions of target data. DINE method \cite{liang2022dine} is employed to generate the crude target model $\phi^{0} $ for the target samples. In the warm-up process, target pseudo-labels $\hat{y}^{src}$ from the source model are used to train a student network to extract features $X$ from the target data to produce prototype labels $\hat{y}^{t}_{\mathds{1}p}$ (pseudo-labels via nearest centroids) and only keep samples that either satisfy the following two conditions: 1) Prototype labels $\hat{y}^{t}_{\mathds{1}p}$ (pseudo-labels via nearest centroids) is consistent with $\hat{y}^{src}_{\mathds{1}}$ (hard label of $\hat{y}^{src}$), and 2) maximum value in $\hat{y}^{src}$ is greater than $\alpha$ (e.g., 0.8). To further remove inaccurate pseudo-labels, we use the intra-class pairwise similarity score to split kept samples into low- and high-confidence data. In addition, samples that do not meet either of the above conditions will be moved to the low-confidence pool. Then, we use selected high-confidence data to train a stronger target network $\phi^{r} $ based on the crude target model $\phi ^{0} $ to improve its performance.

For the incremental process, we first generate prototype labels using the features $X$ from the target network $\phi^{r} $ instead of the student network. Second, we extract high-confidence data from the low-confidence data pool ($L$) for each iteration to the high-confidence data pool ($H$). Subsequently, we train an improved target network  $\phi^{r} $ based on $\phi^{0} $ with samples in the $H$ to relabel the remaining low-confidence data of $L$ in the next epoch. These steps are repeated until the percentage of low-confidence data is less than the threshold $\lambda$ (i.e., $\vert L \vert$/$n_{t}$ $< \lambda$). In contrast to the warm-up process, we only enforce the following criteria to select high-confidence data: 1) pseudo-label consistency between the estimated hard labels and prototype labels, and 2) intra-class similarity between the target samples within the same class.
Given that deep models are generally prone to make overconfident predictions compared to the last iteration, we set an increasingly rigorous screening process for each round (e.g., increasing softmax probability thresholding). The overall framework of the proposed method is summarized in Algorithm \ref{Algorithm1}. In the following, we provide a detailed description of the key components of the framework.

\begin{figure}[tb] %H为当前位置，!htb为忽略美学标准，htbp为浮动图形
\centering %图片居中
\includegraphics[width=0.5\textwidth]{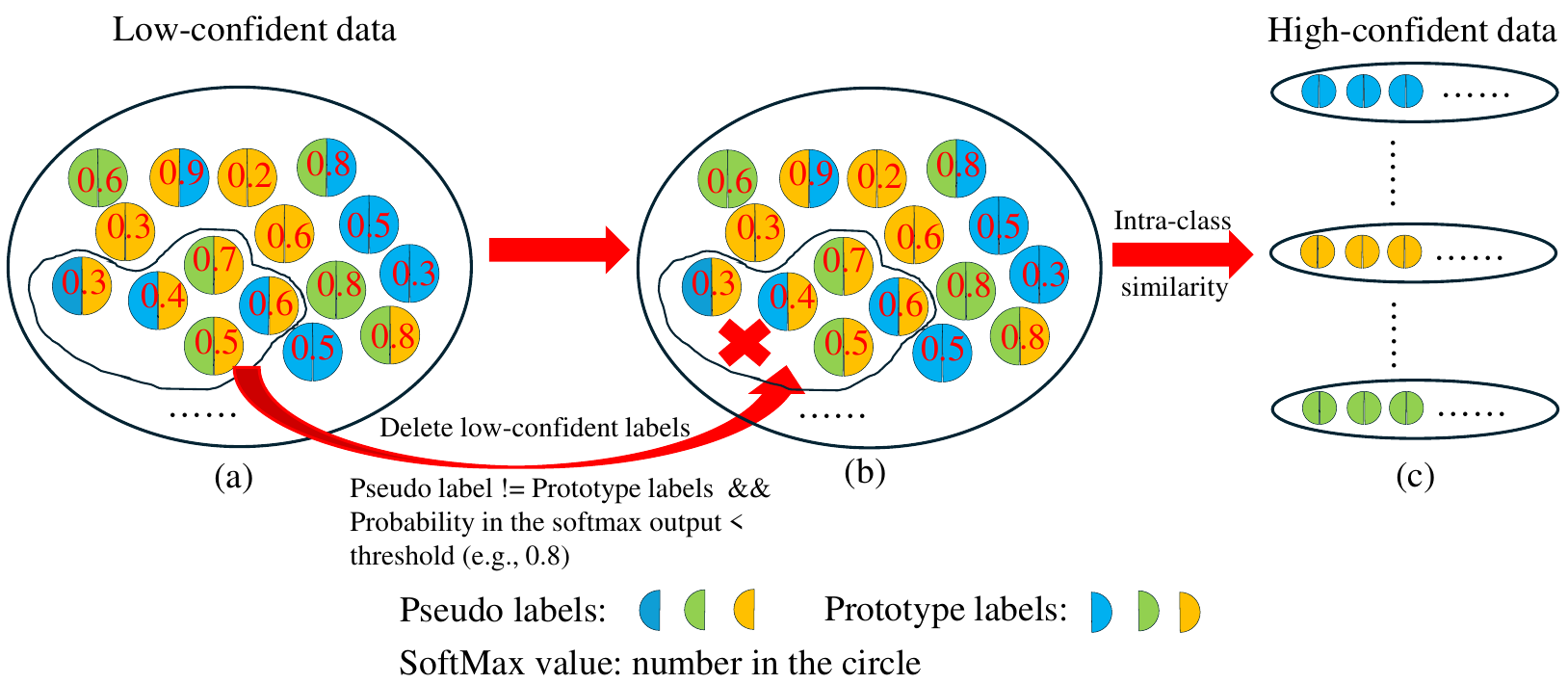} %插入图片，[]中设置图片大小，{}中是图片文件名
\caption{Illustration of strategies used for selecting high-confidence data. %which consists of two steps: (i) Deleting, which deletes
%the samples according to the following rules: pseudo-label not equal prototype labels (pseudo-labels via nearest centroids) and softmax value less thresholding(e.g., 0.8) as shown from Fig. (a) to Fig. (b) (ii) intra-class similarity filtering, which selects samples with high pairwise similarity score as high-confidence data as shown from Fig. (b) to Fig. (c).
} %最终文档中希望显示的图片标题
\label{Figure.2} %用于文内引用的标签
\end{figure}

\subsection{\noindent\textbf{Source model and crude target model generation}}  
For a typical UDA problem, labeled source domain dataset $\mathcal{S} = $ $\left \{ \left (  x^{s}_{i},y^{s}_{i}  \right )  \right \}^{n_{s} }_{i=1} $ with $n_{s}$ samples is given, and unlabeled target domain dataset $\mathcal{T} = $ $\left \{ x^{t} _{i}  \right \} ^{n_{t} } _{i=1} $ lie in different distributions with $n_{t}$ samples. The goal of domain adaptation is to seek the model $f_{t}$ that performs best for predicting target data with the help of the source model $f_{s}$. Considering the nature of the black-box unsupervised domain adaptation task, only the predictions of target data $\hat{y}^{src}$ from the source model are provided without any available source data and any other information from the source model.  
For the scenario of BBUDA, learning starts from obtaining the source model $f_{s}:\mathcal{X} _{s}{\rightarrow}\mathcal{Y} _{s}$ trained with labeled source data by minimizing the following standard cross-entropy loss,
\begin{equation}\label{eq1}
\mathcal{L}_{s}(f_{s}:\mathcal{X} _{s}{\rightarrow}\mathcal{Y}_{s})=-\mathbb{E}_{(x^{s}_{i},y^{s}_{i})\in\mathcal{S}} {\textstyle}\left (q^{s}\right ) \log_{}{f_{s}(x^{s}_{i})}, \ 
\end{equation}
where $q^{s}=\gamma /K+(1-\gamma )y^{s}_{i}$ \cite{muller2019does}  is the smoothed label, $K$ is the number of categories, and $\gamma $ is a smoothing parameter that is empirically set to 0.1.  
After obtaining the source model, we follow  DINE to train the crude target model $\phi^{0} $ for target data with pseudo-labels  $\left \{x^{t},\hat{y}^{src}   \right \}$ from the source model. The difference is that we only use the DINE distilling step without fine-tuning during the training process. Moreover, we train a stronger model $\phi^{r} $ based on the pre-trained model $\phi^{0} $ with high-confidence data selected via the method introduced in the next section.  

\subsection{\noindent\textbf{Selection of high-confidence data}}  
Mislabeling, which leads to substantial cumulative errors in the iterative training process, is commonly observed in pseudolabeling. To address this, we draw inspiration from the simple intuition that a stronger target network can be more easily trained once the high-confidence data can be effectively selected to serve as a reliable input. Hence, to achieve a stronger model, we increase the amount of high-confidence data and corrected the inaccurate pseudo-labels of low-confidence data incrementally.
Thus, we propose the following three-step pseudo-label selection strategy to search for high-confidence data. For ease of understanding, we present an illustration in Fig. \ref{Figure.2} to show the strategies.
\subsubsection{\noindent\textbf{Thresholding of softmax probabilities}}
For most deep networks, the softmax function produces a probability distribution over $k$ known class labels. It is trivial to realize that classes with high probability are believed to have high confidence. Hence, we first introduce a thresholding procedure for the softmax outputs, which rejects low-confidence data while maintaining high-confidence data. In particular, in the warm-up process, if $\hat{y}^{src}$ is greater than $\alpha$, we regard it as high-confidence data, otherwise low-confidence data. %In the incremental process, we set a low threshold $\eta$ starting with 0.1 to discard prototype labels with softmax outputs below 0.1 to generate more accurate labels for three tasks. Furthermore, we set an increasingly rigorous screening criterion by increasing $\eta$ iteratively via $\eta = 0.1 + 0.1*cyclenum$ to obtain high-confidence data, where $cyclenum$ is the incremental iteration number of Pseudo-Labeling. 

\subsubsection{\noindent\textbf{Prototype labels}}
We apply the pseudo-labeling strategy described in SHOT \cite{liang2020we} to generate prototype labels (pseudo-labels via nearest centroids). However, this requires feature representations that are inaccessible to BBUDA. Hence, we introduce a student network $f_{st}:\mathcal{X} ^{t}{\rightarrow}\mathcal{Y} ^{t}$ to learn the structural information from the source model via teacher-student knowledge distillation.  In this stage, we train the student network with pseudo-labels from the source model by minimizing the following Kullback-Leibler (KL) divergence loss, aiming to extract target features,
\begin{equation}\label{eq2}
Lst_{kd} = \mathbb{E}_{x^{t}\in\mathcal{T} }D_{kl} (f_{s}(x^{t} ) \left |  \right | f_{st}(x^{t} )), \ 
\end{equation}
We then attain the centroid for each class in the target domain, similar to weighted k-means clustering, as follows:
\begin{equation}\label{eq3}
c^{(0)}_{k}  =\frac{ {\textstyle \sum_{x^{t}{{\scriptsize {\in}} }\mathcal{T}}} f_{t}(x^{t})g_{t}(x^{t})}{ {\textstyle \sum_{x^{t}{{\scriptsize {\in}} }\mathcal{T}} f_{t}(x^{t})} }, \ 
\end{equation}
where $f_{t}(x^{t})$ denotes the output of the softmax layer and $g_{t}(x^{t})$ are features extracted before the classifier module. Then,
\begin{equation}\label{eq4}
\hat{y}^{t}_{p} = \mathrm {arg}{\ } \mathrm{min} D_{f}(g_{t}(x^{t}),c^{(0)}_{k}), \ 
\end{equation}
where $D_{f}(g_{t}(x^{t}),c^{(0)}_{k})$ measures the cosine distance between $g_{t}(x^{t})$ and $c^{(0)}_{k}$. 
Then, update the centroids and labels in Eq. \ref{eq5} for one round:
\begin{equation}\label{eq5}
\begin{aligned}
c^{(1)}_{k} &=\frac{ {\textstyle \sum_{x^{t}{{\scriptsize {\in}} }\mathcal{T}}} \mathds{1}(\hat{y}^{t}_{p}=k)   g_{t}(x^{t})}{ {\textstyle \sum_{x^{t}{{\scriptsize {\in}} }\mathcal{T}} \mathds{1}(\hat{y}^{t}_{p}=k)} }, \\
\hat{y}^{t}_{\mathds{1}p} &= \mathrm {arg}{\ } \mathrm{min} D_{f}(g_{t}(x^{t}),c^{(1)}_{k}), \ 
\end{aligned}
\end{equation}
The centroid for each class closest to the feature is used as the prototype label. Unlike \cite{liang2022dine,liang2020we} that directly utilize the closest centroid as the prototype label without selection, we take the relationship between the smallest distance and the second-smallest distance into consideration. Specifically,
\begin{equation}\label{eq6}
d_{t}  =\frac{ \mathrm{2ndmin} D_{f}(g_{t}(x^{t}),c^{(1)}_{k})- \mathrm{min} D_{f}(g_{t}(x^{t}),c^{(1)}_{k})}{\mathrm{min} D_{f}(g_{t}(x^{t}),c^{(1)}_{k})},  \
\end{equation}
%where $\mathrm{min} D_{f}(g_{t}(x^{t}),c_{k}^{(0)})$ and $\mathrm{2ndmin} D_{f}(g_{t}(x^{t}),c_{k}^{(0)})$ are the centroid of the category at the minimum and second smallest distance from the feature respectively. 
To select reliable prototype labels $\hat{y}_{\mathds{1}p}^{t}$, we set a threshold $\beta$ to keep those whose minimum distance is significantly closer than the second-smallest distance. In particular, if the hard label of $\hat{y}^{t}_{\mathds{1}}$ is the same as $\hat{y}_{\mathds{1}p}^{t}$ and $d_{t} >\beta$, the pseudo-labels are regarded as high-confidence data. 
\subsubsection{\noindent\textbf{Intra-class similarity}}

To further improve the accuracy of pseudo-labels, we utilize intra-class similarity \cite{wang2023improving} to select pseudo-labels with high conﬁdence. The motivation behind this is that samples with the same target class labels should have high intra-class similarity scores. By contrast, samples classified into different target classes should have low intra-class similarity scores. Hence, to remove low-confidence data, we calculate a pairwise similarity matrix $S_{i,j}^{k}$ of the target samples for each pseudo-class ($k =$1, 2, 3, \dots, $C_{y}$) following \cite{wang2023improving} via
\begin{equation}\label{eq7} 
S_{i,j}^{k} =  \frac{\left \langle X_{i}^{tk},X_{j}^{tk}  \right \rangle}{\left \| X_{i}^{tk} \right \| \left \| X_{j}^{tk} \right \| }, \
\end{equation}
where $S_{i,j}^{k}$ is the cosine similarity between the $i$-th sample and $j$-th sample of the ${k}$-th pseudo class, $X_{i}^{tk}$ and $X_{j}^{tk}$ are feature vector of $i$-th sample and $j$-th sample of the ${k}$-th pseudo class. $\left \langle X_{i}^{tk}, X_{j}^{tk}  \right \rangle$ is the inner product of two feature vector $ X_{i}^{tk}, X_{j}^{tk}$, and $\left \| X_{i}^{tk} \right \|, \left \| X_{j}^{tk} \right \|$ are the  L2 normalization operation of $ X_{i}^{tk}, X_{j}^{tk}$. To select high-confidence pseudo-labels, we set a similarity threshold $\delta$,
\begin{equation}\label{eq8}
M_{i,j}^{k} = \begin{cases}1, \ \ S_{i,j}^{k}>\delta
 \\0,\ \ \mathrm{otherwise}, \

\end{cases}
\end{equation}

\begin{algorithm}[tb]
\caption{Incremental Pseudo-Labeling}
\label{Algorithm1}
\begin{algorithmic}[1]
\REQUIRE{Labeled source samples, $\mathcal{S} = $ $\left \{ \left (  x^{s}_{i},y^{s}_{i}  \right )\right\}^{n_{s} }_{i=1} $, Unlabeled target samples, $\mathcal{T} = $ $\left \{ x^{t} _{i}\right\} ^{n_{t} } _{i=1} $};
\ENSURE{Target model $f_{t}$};
\STATE \textbf{1. Training Source Model and Crude Target Model.}
\STATE  Train source model $f_{s}$ with $\mathcal{S}$ by minimizing Eq. \ref{eq1};
\STATE  Obtain the crude target model $\phi^{0} $ by DINE \cite{liang2022dine};
\STATE  Initialize Pseudo-labels  $\left \{x^{t},\hat{y}^{t}   \right \}$ by $\hat{y}^{t}=\hat{y}^{src}$;
\STATE  \textbf{2. Selecting high-confidence samples.}
\STATE Train student network with pseudo-labels to obtain the features to generate prototype labels  $\left \{x^{t},\hat{y}^{t}_{\mathds{1}p}   \right \}$;
\STATE Initialize low-confidence data pool, $L = \left\{x^{t},\hat{y}^{t}\right\}$;
\STATE Initialize high-confidence data pool, $H = \emptyset$;
\STATE \textbf{2.1) Warm-up process.}
\FOR{\textbf{each} $x^t \in L$}
\IF{($\hat{y}^{t}_{\mathds{1}p}==\hat{y}^{t}_{\mathds{1}}$ or $\hat{y}^{t}>\alpha$) and ( $ts^{k} > \theta$)}
\STATE Treat $x^t$ as a high-confidence sample and move $x^t$ from $L$ to $H$;
\ENDIF
\ENDFOR
\STATE $r=1$;
\STATE Train $\phi^{r} $ with $H$ based on $\phi^{0} $ by minimizing Eq. \ref{eq15};
\STATE \textbf{2.2) Incremental process.}
\WHILE{$\vert L \vert$/$n_{t}$ $ \geq \lambda$ }
\STATE $r=r+1$;
\FOR{\textbf{each} $x^t \in L$}
\IF{ ($\hat{y}^{t}_{\mathds{1}p}$ == $\hat{y}^{t}_{\mathds{1}}$) and ($ts^{k} > \theta$) }
\STATE Move $x^t$ from $L$ to $H$;
\ENDIF
\ENDFOR
\STATE Use $H$ to train $\phi^{r} $ based on $\phi^{0} $ by minimizing Eq. \ref{eq15};
\IF{$\vert L \vert$/$n_{t}$ $< \lambda$}
\STATE Fine-tune $\phi^{r}$ by minimizing Eq. \ref{eq13};
\RETURN $\phi^{r} $ as $f_{t}$;
\ENDIF
\ENDWHILE
\end{algorithmic}
\end{algorithm}

\begin{equation}\label{eq9}
ts_{i}^{k} =\frac{   {\textstyle \sum_{j\in C_{k}}M_{i,j}^{k}}}{n^{k}}, \
\end{equation}
where $\delta$ is set to 0.6 in the experiment, $ C_{k}$ is the ${k}$-th pseudo class, $ts_{i}^{k}$ denotes the percentage of the pseudo class $ C_{k}$ similar to $i$-th sample of the same class, and $n^{k}$ is the number of samples for $ C_{k}$. We set the similarity threshold $ \theta$ to 0.3 for all tasks. If $ts_{i}^{k}>$ $ \theta$, we regard the pseudo-labels as high-confidence data. Otherwise, we regard them as low-confidence data.

\subsection{\noindent\textbf{Overall Objectives}}  
Our training objective for the black-box unsupervised domain adaptation consists of several loss functions. First, we apply teacher-student knowledge distillation \cite{hinton2015distilling} to our method to transfer knowledge from the source to the target model,
\begin{equation}\label{eq10}
L_{kd} = \mathbb{E}_{x^{t}\in\mathcal{T}}D_{kl} (f_{s}(x^{t} ) \left |  \right | f_{t}(x^{t} )), \
\end{equation}
where $D_{kl}$ denotes the Kullback-Leibler (KL) divergence
loss.
Second, we apply the widely-used mutual information objective loss to keep target outputs similar to one-hot encoding and categories globally diverse following \cite{liang2020we},
\begin{equation}\label{eq11}
L_{ent}   =- \textstyle \sum_{x^{t}{{\scriptsize {\in}} }\mathcal{T}} f_{t}(x^{t})\log_{}{f_{t}(x^{t})}, \
\end{equation}
\begin{equation}\label{eq12}
L_{div}   =- \textstyle \sum_{k=1}^{K}  \hat{p} _{k} \log_{}{\hat{p} _{k}}, \
\end{equation}
\begin{equation}\label{eq13}
L_{im}   =L_{ent}+L_{div}, \
\end{equation}
where $K$ is the number of classes and $\hat{p} _{k}$ is the mean output embedding of the ${k}$-th pseudo class.
Third, we utilize MixUp and interpolation consistency training for data augmentation to improve the generalization ability following \cite{liang2022dine},
\begin{equation}\label{eq14}
\begin{aligned}
L_{mix}   & = \mathbb{E}_{x^{t}_{i},x^{t}_{j}\in \mathcal{T}}\mathbb{E}_{\eta \in Beta(\omega ,\omega )} \\
& l_{kl}(\mathrm{Mix}_{\eta } ({f_{t}}' (x^{t}_{i}),{f_{t}}'(x^{t}_{j})),f_{t}(\mathrm{Mix}_{\eta }(x^{t}_{i},x^{t}_{j}))), \
\end{aligned}
\end{equation}
where $l_{kl}$ denotes the Kullback-Leibler divergence loss, and $\mathrm{Mix}_{\eta}(a,b)=\eta  a +(1-\eta ) b$ denotes the MixUp operation. $\eta$ is sampled from a Beta distribution, and $\omega$ is the hyper-parameter empirically set to 0.3.
Based on the objectives introduced above, we conclude the full objective for black-box unsupervised domain adaptation as follows,
\begin{equation}\label{eq15}
L_{t}   =L_{kd}+L_{im}+L_{mix}, \
\end{equation}

After the incremental process, we finally employ mutual information maximization in Eq. \ref{eq13} to finetune the target model to improve its generalization ability.

\section{Experiment}
\subsection{\noindent\textbf{Setup}}  
\noindent\textbf{Datasets.} We evaluate the model performance via three popular benchmark datasets on object recognition:
\begin{itemize}
    \item \textbf{Office} \cite{saenko2010adapting} is a small-size dataset containing three different domains, A (Amazon), W (Webcam), and D (DSLR), with 31 classes and 4,652 images. 

    \item \textbf{Office-Home} \cite{venkateswara2017deep} is a medium-sized benchmark consisting of four distinct domains: Ar (Art), Cl (Clipart), Pr (Product), and Re (Real World), with 65 categories and 15,500 images. 

    \item \textbf{VisDA-C} \cite{peng2017visda} is a challenging large-size dataset that focuses on the synthesis-to-real object recognition task with 152 thousand synthetic images in the source domain and 55 thousand real object images in the target domain. 
\end{itemize}

\noindent\textbf{Implementation Details.}   Three independent runs are performed, and the average classification accuracy is recorded. Following \cite{liang2022dine}, we employ two different backbones: ResNet-50 for the Office and Office-Home dataset and ResNet-101 \cite{he2016deep} for VisDA-C. For a fair comparison, we generate the crude target model using the same models as the backbone network. Training parameters are learning rate ($1e^{-2}$), momentum (0.9), weight decay ($1e^{-3}$), bottleneck size (256), and batch size (64) following the protocols introduced in DINE \cite{liang2022dine}. We train 30 epochs to learn structural knowledge from the source model for the Office and Office-Home datasets and 10 epochs for VisDA-C. For the hyperparameters in our method, $\alpha=0.8$, $\beta=0.3$, $\lambda=0.1$ for Office; $\alpha=0.6$, $\beta=0.2$, $\lambda=0.2$ for Office-Home; and $\alpha=0.7$, $\beta=0.2$, $\lambda=0.25$ for VisDA-C are used. For the fine-tuning step, we train 30 epochs for the Office and Office-Home datasets and 10 epochs for VisDA-C.

% \textbf{Image-CLEF} \cite{long2017deep} consists of three domains, each with 12 classes and 600 images. \textbf{Office-Caltech} \cite{gong2012geodesic} comprises four domains with ten shared categories between Office and Caltech-256. 
\noindent\textbf{Baselines.} To be consistent with \cite{liang2022dine}, we compare our method against single-source adaptation models: NLL-KL \cite{zhang2021unsupervised}, NLL-OT \cite{asano2019self}, SD-SHOT \cite{liang2020we}, HD-SHOT \cite{liang2020we}, DINE \cite{liang2022dine}, and DINE-full \cite{liang2022dine}.

\begin{table*}[ht]
\caption{Accuracies (\%) on Office for single-source closed-set UDA.}
\label{tab:t1}
\centering
\begin{tabular*}{\hsize}{@{}@{\extracolsep{\fill}}lccccccc@{}}
\toprule
Method      & A→D  & A→W  & D→A  & D→W  & W→A  & W→D  & Avg. \\ \hline
No Adapt. \cite{liang2022dine} & 79.9 & 76.6 & 56.4 & 92.8 & 60.9 & 98.5 & 77.5 \\
NLL-OT \cite{asano2019self}      & 88.8 & 85.5 & 64.6 & 95.1 & 66.7 & \textbf{98.7} & 83.2 \\
NLL-KL \cite{zhang2021unsupervised}     & 89.4 & 86.8 & 65.1 & 94.8 & 67.1 & \textbf{98.7} & 83.6 \\
HD-SHOT \cite{liang2020we}    & 86.5 & 83.1 & 66.1 & 95.1 & 68.9 & 98.1 & 83.0   \\
SD-SHOT \cite{liang2020we}    & 89.2 & 83.7 & 67.9 & 95.3 & 71.1 & 97.1 & 84.1 \\
DINE \cite{liang2022dine}      & 91.6 & 86.8 & 72.2 & 96.2 & 73.3 & 98.6 & 86.4 \\
DINE (full) \cite{liang2022dine}  & 91.7 & 87.5 & 72.9 & \textbf{96.3} & 73.7 & 98.5 & 86.7 \\
Ours        & \textbf{94.6} & \textbf{88.1} & \textbf{74.0} & 95.8 & \textbf{75.3} & 98.3 & \textbf{87.7} \\
\bottomrule
\end{tabular*}

\end{table*}

\begin{table*}[ht]
\caption{Accuracies (\%) on Office-Home for single-source closed-set UDA.}
\label{tab:t2}
\centering
\scalebox{1}{
\begin{tabular*}{\hsize}{@{}@{\extracolsep{\fill}}lccccccccccccc@{}}
\toprule
Method     & Ar→Cl & Ar→Pr & Ar→Re & Cl→Ar & Cl→Pr & Cl→Re & Pr→Ar & Pr→Cl & Pr→Re & Re→Ar & Re→Cl & Re→Pr & Avg. \\ \hline
No Adapt. \cite{liang2022dine}        & 44.1  & 66.9  & 74.2  & 54.5  & 63.3  & 66.1  & 52.8  & 41.2  & 73.2  & 66.1  & 46.7  & 77.5  & 60.6 \\
NLL-OT \cite{asano2019self}     & 49.1  & 71.7  & 77.3  & 60.2  & 68.7  & 73.1  & 57.0    & 46.5  & 76.8  & 67.1  & 52.3  & 79.5  & 64.9 \\
NLL-KL \cite{zhang2021unsupervised}     & 49.0    & 71.5  & 77.1  & 59.0    & 68.7  & 72.9  & 56.4  & 46.9  & 76.6  & 66.2  & 52.3  & 79.1  & 64.6 \\
HD-SHOT \cite{liang2020we}    & 48.6  & 72.8  & 77.0    & 60.7  & 70.0    & 73.2  & 56.6  & 47.0    & 76.7  & 67.5  & 52.6  & 80.2  & 65.3 \\
SD-SHOT \cite{liang2020we}   & 50.1  & 75.0    & 78.8  & 63.2  & 72.9  & 76.4  & 60.0    & 48.0    & 79.4  & 69.2  & 54.2  & 81.6  & 67.4 \\
DINE \cite{liang2022dine}       & 52.2  & \textbf{78.4}  & 81.3  & 65.3  & 76.6  & 78.7  & 62.7  & 49.6  & 82.2  & 69.8  & 55.8  & 84.2  & 69.7 \\
DINE (full) \cite{liang2022dine} & 54.2  & 77.9  & 81.6  & 65.9  & 77.7  & 79.9  & 64.1  & 50.5  & 82.1  & 71.1  & 58.0   & 84.3  & 70.6 \\
Ours       & \textbf{54.9}  & 77.6  & \textbf{81.7}  & \textbf{68.3}  & \textbf{78.4}  & \textbf{80.5}  & \textbf{65.7}  & \textbf{52.6}  & \textbf{82.6}  & \textbf{71.5}  & \textbf{58.2}  & \textbf{84.8}  & \textbf{71.4} \\
\bottomrule
\end{tabular*}
}
\end{table*}

\begin{table*}[]
\caption{Accuracies (\%) on n VisDA-C for single-source closed-set UDA.}
\label{tab:t3}
\centering
\scalebox{1}{
\begin{tabular*}{\hsize}{@{}@{\extracolsep{\fill}}lccccccccccccc@{}}
\toprule
Method  & plane & bcycl & bus  & car  & horse & knife & mcycle & person & plant & sktbrd & train & truck & Per-class \\ \hline
No Adapt. \cite{liang2022dine}       & 64.3  & 24.6  & 47.9 & \textbf{75.3} & 69.6  & 8.5   & 79.0     & 31.6   & 64.4  & 31.0     & 81.4  & 9.2   & 48.9      \\
NLL-OT \cite{asano2019self}  & 82.6  & 84.1  & 76.2 & 44.8 & 90.8  & 39.1  & 76.7   & 72.0     & 82.6  & 81.2   & 82.7  & 50.6  & 72.0        \\
NLL-KL \cite{zhang2021unsupervised}   & 82.7  & 83.4  & 76.7 & 44.9 & 90.9  & 38.5  & 78.4   & 71.6   & 82.4  & 80.3   & 82.9  & 50.4  & 71.9      \\
HD-SHOT \cite{liang2020we}  & 75.8  & 85.8  & 78.0   & 43.1 & 92.0    & \textbf{41.0}    & 79.9   & 78.1   & 84.2  & 86.4   & 81.0    & \textbf{65.5}  & 74.2      \\
SD-SHOT \cite{liang2020we}  & 79.1  & 85.8  & 77.2 & 43.4 & 91.6  & \textbf{41.0}    & 80.0     & 78.3   & 84.7  & 86.8   & 81.1  & 65.1  & 74.5      \\
DINE \cite{liang2022dine}     & 81.4  & \textbf{86.7}  & 77.9 & 55.1 & 92.2  & 34.6  & 80.8   & 79.9   & 87.3  & 87.9   & 84.3  & 58.7  & 75.6      \\
DINE (full) \cite{liang2022dine}     & \textbf{95.3}  & 85.9 & \textbf{80.1} & 53.4 & 93.0    & 37.7  & 80.7   & 79.2   & 86.3  & \textbf{89.9}   & 85.7  & 60.4  & 77.3      \\
Ours    & 88.4  & 85.3  & 78.5 & 62.6 & \textbf{94.8}  & 38.1  & \textbf{88.2}   & \textbf{81.9}   & \textbf{89.5}  & 88.8   & \textbf{87.2}  & 59.3  & \textbf{78.6} \\
\bottomrule
\end{tabular*}
}
\end{table*}

\subsection{\noindent\textbf{Results}}  
We evaluated our method for domain adaptation using the following three object recognition benchmarks: Office, Office-Home, and VisDA-C. The results are reported in Tab. \ref{tab:t1}, Tab. \ref{tab:t2}, and Tab. \ref{tab:t3}, respectively. It can be observed that our method achieves the best mean accuracy for the three tasks and outperforms prior methods. For the small-scale dataset Office, our method yields improvements of 1.0\% in terms of the average accuracy, as shown in Tab. \ref{tab:t1} and performs the best among 4 out of 6 tasks in Office. Specifically, our method produces lower accuracy for A→D and W→D. This can be attributed to the fact that our model incrementally increases the high-confidence data, which makes it difficult to further improve the positive samples with saturated accuracies initially.
For the medium-sized Office-Home dataset listed in Tab. \ref{tab:t2}, our method increases the average accuracy from 70.6\%  to 71.4\% and performs the best among 11 out of 12 tasks. 
The superior performance is largely attributed to the incremental pseudo-labeling mechanism, which gradually increases the amount of high-confidence data, thereby promoting the ability of the network to learn quality feature representations. For the large-scale VisDA-C dataset in Tab. \ref{tab:t3}, our method surpasses DINE by 1.3\% in terms of average accuracy. It is interesting to note that the accuracy improved by our method on VisDA-C varies greatly from -6.9\% to 7.5\% for different classes, which is mostly due to the imbalanced class distributions of high-confidence data. 
Overall, the results demonstrate that the proposed method can effectively improve the classification performance of the target model.

\subsection{\noindent\textbf{Analysis}}  
\noindent\textbf{Ablation Study.} First, we studied the contributions of the three losses in our method. The results for the office dataset are presented in Tab. \ref{tab:t4}, where four different cases are listed according to the composition of different losses. It can be confirmed that the combination of  $L_{kd}$\&$L_{im}$ and $L_{kd}$\&$L_{im}$\&$L_{mix}$ outperforms DINE and DINE (full) by 0.6\% and 1.0\%, respectively, in terms of average accuracy. Notably, using only $L_{kd}$\&$L_{mix}$ produces a lower accuracy, and $L_{kd}$ performs competitively with DINE. This indicates that each component in our training objective has a positive impact on improving the performance of the target model. Besides, it demonstrates that
using only $L_{kd}$ cannot ensure the efficacy of learning feature representations from high-confidence data.

\begin{figure*}[]
\centering
\subfloat[] {\includegraphics[width=0.3\linewidth,scale=0.5]{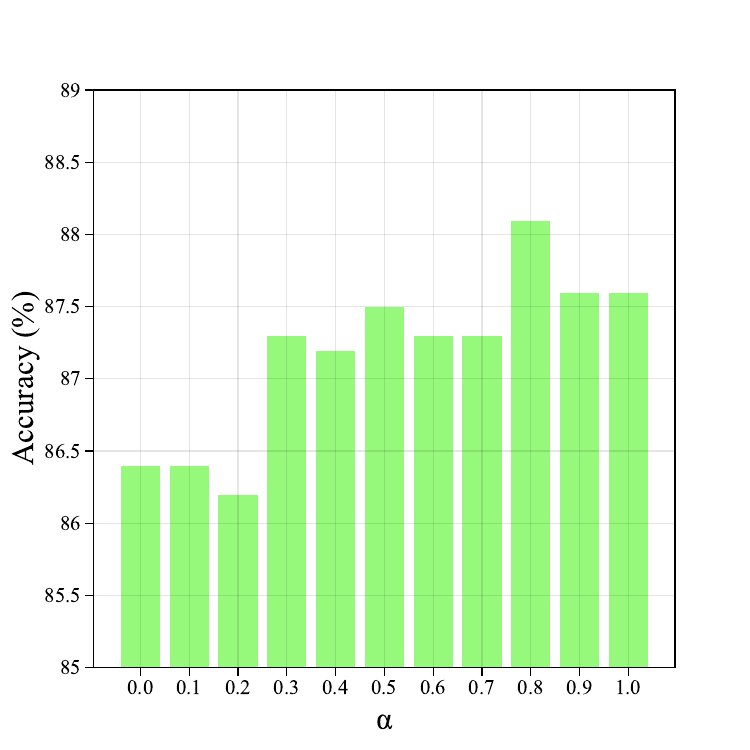}}\hspace{0.5em}%
\subfloat[]
{\includegraphics[width=0.3\linewidth,scale=0.5]{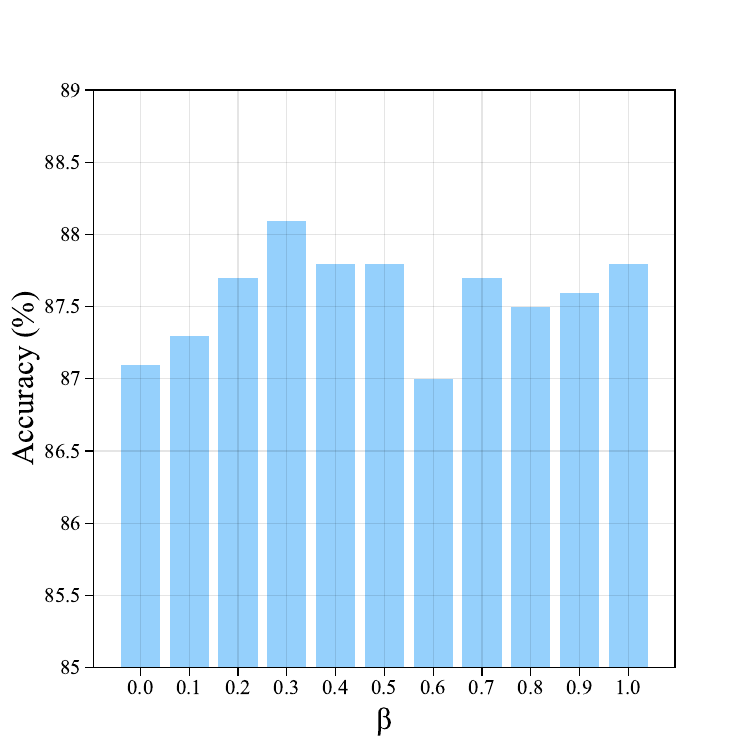}}\hspace{0.5em}%
\subfloat[]
{\includegraphics[width=0.3\linewidth,scale=0.5]{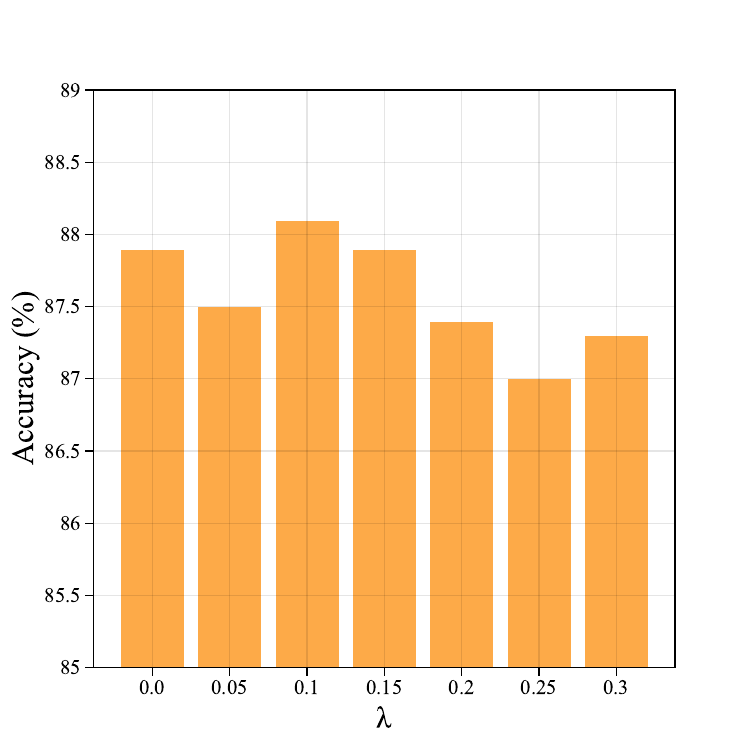}}\hspace{0.5em}%
\caption{Parameter Analysis}
\label{fig:5}
\end{figure*}

\textbf{Parameter Study.}
We select the task A$\to$W from Office to study the parameter sensitivity of $\alpha$ (thresholding of softmax probabilities), $\beta$ (the significant differences between the smallest and the second-smallest distance), $\lambda$ (the proportion of low-confidence data) in Fig. \ref{fig:5}(a-c),
where $\alpha$ and $\beta$ are from 0.0 to 1.0 with interval 0.1, and $\lambda$ is from 0.0 to 0.3 with interval 0.05. We can see that the larger the $\alpha$, the better the model performs. We assume this is because alpha directly controls the quality of high-confidence data, and a greater alpha maintains more reliable pseudo-labels. For parameter $\beta$, we can see that the model’s performance increases in the early stage but decreases gradually in the test range, as shown in Fig. \ref{fig:5}(b). We assume that this is a consequence of $\beta$ rejecting more incorrect labels at an early stage instead of correct labels, whereas high $\beta$ discards part of the high-confidence data. For the parameter $\lambda$, the smaller the $\lambda$, the more high-confidence data we obtain. The overall performance degrades at the later training stage, which results from a decrease in high-confidence data. Note that parameter $\alpha$ has a higher sensitivity than the other parameters during optimization in the range of study, which can reject low-confidence classes while maintaining the high-confidence data effectively.

\begin{table}[ht]
\setlength\tabcolsep{3pt}
\caption{Ablation Study of Losses on Selected Tasks}
\label{tab:t4}
\centering
\scalebox{1}{
\begin{tabular}{lccccccccccccc}

\toprule
Methods       & A→D  & A→W  & D→A  & D→W  & W→A  & W→D  & Avg. \\
DINE          & 91.6 & 86.8 & 72.2 & 96.2 & 73.3 & \textbf{98.6} & 86.4 \\
DINE (full)    & 91.7 & 87.5 & 72.9 & \textbf{96.3} & 73.7 & 98.5 & 86.7 \\ \hline
$L_{kd}$          & 92.8 & 86.4 & 72.2 & 95.2 & 73.5 & 98.1 & 86.4 \\
$L_{kd}$\&$L_{im}$      & 93.9 & 87.2 & 73.9 & 95.5 & 74.8 & 98.4 & 87.3 \\
$L_{kd}$\&$L_{mix}$     & 92.4 & 86.0 & 71.5 & 95.3 & 71.9 & 98.1 & 85.9 \\
$L_{kd}$\&$L_{im}$\&$L_{mix}$ & \textbf{94.6} & \textbf{88.1} & \textbf{74.0} & 95.8 & \textbf{75.3} & 98.3 & \textbf{87.7} \\
\toprule
\end{tabular}
}
\end{table}

\section{Conclusion}
In this study, we propose a simple yet effective incremental pseudo-labeling method to improve the accuracy of pseudo-labels for black-box unsupervised domain adaptation. Speciﬁcally, we ﬁrst select high-confidence data via thresholding of softmax probabilities and prototype labels and then discard data with low intra-class similarity. We train a stronger target network with high-confidence data to correct the pseudo-labels of low-confidence data. Our method regards pseudo-labels of the target data from the source model as a low-confidence data pool and incrementally selects high-confidence data from it to the high-confidence data pool. Furthermore, the high-confidence data pool grows, and the low-confidence data pool decreases iteratively. Experimental results on three datasets validate that our method can improve the accuracy of the pseudo-labels and further achieve better generalization performance than conventional UDA baselines. 

Although incremental pseudo-labeling shows satisfactory accuracy for black-box unsupervised domain adaptation, the improved performance is affected heavily by the imbalance of high-confidence data, which produces serious bias for the majority class. This could be resolved by applying weighted cross-entropy loss to assign more weights to the minority classes. We would like to explore this issue in the future.

\ifCLASSOPTIONcaptionsoff
  \newpage
\fi



\bibliographystyle{IEEEtran}
\bibliography{ref}
\end{document}